\documentclass[conference]{IEEEtran}
\usepackage{amsmath,graphicx}
\usepackage{verbatim}
\usepackage{mathrsfs}
\usepackage{amsopn}
\usepackage{cite}
\usepackage{amsthm}
\usepackage{epsfig,psfrag}
\usepackage{amsbsy,amsfonts,amssymb}
\usepackage{algorithm}
\usepackage{algorithmic}
\usepackage{graphics}
\usepackage{epsfig}
\usepackage{epstopdf}
\usepackage{xcolor}
\usepackage{array}
\usepackage{arydshln}
\usepackage{color}
\usepackage{xcolor}
\usepackage{nicefrac}
\usepackage{hyperref}
\usepackage{cleveref}
\usepackage{tabularx}
\usepackage{multirow}
\usepackage{booktabs}
\usepackage{xfrac}
\usepackage{graphicx}
\usepackage{subfig}
\usepackage{textcomp}
\graphicspath{{figures/}}
\newcommand{\bb}{\boldsymbol}
\IEEEoverridecommandlockouts
\def\BibTeX{{\rm B\kern-.05em{\sc i\kern-.025em b}\kern-.08em
    T\kern-.1667em\lower.7ex\hbox{E}\kern-.125emX}}
\begin{document}

\title{Convolutional Simultaneous Sparse Approximation with Applications to RGB-NIR Image Fusion
}

\author{\IEEEauthorblockN{Farshad G. Veshki}
\IEEEauthorblockA{\textit{Dept. Signal Processing and Acoustics} \\
\textit{Aalto University}\\
Espoo, Finland \\
farshad.ghorbaniveshki@aalto.fi}
\and
\IEEEauthorblockN{Sergiy A. Vorobyov}
\IEEEauthorblockA{\textit{Dept. Signal Processing and Acoustics} \\
\textit{Aalto University}\\
Espoo, Finland \\
sergiy.vorobyov@aalto.fi}}

\maketitle

\begin{abstract}
Simultaneous sparse approximation (SSA) seeks to represent a set of dependent signals using sparse vectors with identical supports. The SSA model has been used in various signal and image processing applications involving multiple correlated input signals. In this paper, we propose algorithms for convolutional SSA (CSSA) based on the alternating direction method of multipliers. Specifically, we address the CSSA problem with different sparsity structures and the convolutional feature learning problem in multimodal data/signals based on the SSA model. We evaluate the proposed algorithms by applying them to multimodal and multifocus image fusion problems. 
\end{abstract}


\section{Introduction}
Simultaneous sparse approximation (SSA) aims to reconstruct multiple input signals using sparse representations (SRs) with identical supports, i.e., using different linear combinations of the same subset of atoms in a dictionary~\cite{ssa_greedy,ssa_convex}. The SSA problem can be written as follows
\begin{equation}
\begin{aligned}
\underset{\{\bb{x}_n\}_{n=1}^N}{\mathrm{minimize}} \; & \sum_{n=1}^N\left( \frac{1}{2}\left\|\bb{D}\bb{x}_n -\bb{s}_n\right\|_2^2 +\lambda \left\|\bb{x}_n\right\|_{0}\right)\\ 
\text{s.t.} \quad &\mathrm{Supp}\big(\bb{x}_{l}\big)=\mathrm{Supp}\big(\bb{x}_{m}\big),\quad l,m=1,\dots,N, \label{eq: SSA}
\end{aligned}
\end{equation}
where $\bb{D}$, $\{\bb{x}_n\}_{n=1}^N$ and $\{\bb{s}_n\}_{n=1}^N$ represent the dictionary, the SRs with identical supports, and the input signals, respectively. Moreover, $\lambda>0$ is the sparsity regularization parameter, $\|\cdot\|_2$ is the Euclidean norm, $\|\cdot\|_0$ is an operator that counts the nonzero entries of a vector, and $\mathrm{Supp}(\cdot)$ denotes the support of an array. The simultaneous sparsity model has been used in a wide range of signal and image processing applications involving multiple dependent input signals. For example, multi measurement vectors (MMV) problems~\cite{MMV1,MMV2}, image fusion~\cite{Fusion_SSA,Fusion_CFL}, anomaly detection~\cite{Anomaly}, and blind source separation~\cite{BSS}. 

Problem \eqref{eq: SSA} is non-convex and, in general, intractable in polynomial time. A common approach for addressing the SSA problem is convex relaxation using mixed-norms~\cite{ssa_convex,baysian_ssa}. For a matrix $\bb{A}\in\mathbb{R}^{R\times C}$, the mixed $\ell_{p,q}$-norm, $p,q\geq1$, is defined as
\begin{equation*}
    \|\bb{A}\|_{p,q} = \left(\sum_{r=1}^R \|\bb{A}(r,\cdot)\|_p^q \right)^{\frac{1}{q}}
\end{equation*}
where $\bb{A}(r,\cdot)$ is the $r$th row of $\bb{A}$, and $\|\cdot\|_p$ denotes the p-norm of a vector. For example, the $\ell_{2,1}$ and the $\ell_{\infty,1}$-norms have been used for addressing the SSA problem in \cite{Duarte2016} and \cite{ssa_convex}, respectively. An unconstrained convex relaxation of \eqref{eq: SSA} using the $\ell_{2,1}$-norm can be written as 
\begin{equation}
\underset{\bb{X}}{\mathrm{minimize}} \; \frac{1}{2}\sum_{n=1}^N \left\|\bb{D}\bb{x}_n -\bb{s}_n\right\|_2^2 +\lambda \left\|\bb{X}\right\|_{2,1}, \label{eq: SSA convex 12}
\end{equation}
where $\bb{X} = [\bb{x}_1 \cdots \bb{x}_N]$. Solving \eqref{eq: SSA convex 12} entails minimizing the $\ell_2$-norm of the rows (enforcing dense rows) and the sum of the $\ell_2$-norms of the rows (promoting all-zero rows) of $\bb{X}$. Thus, the resulting $\bb{X}$ is expected to be mostly zeros with only few non-zero and dense rows. This structure is referred to as \textit{row-sparse} structure. A \textit{row-sparse} structure with sparse rows can be enforced by embedding an additional $\ell_1$-norm regularization term in the objective function of~\eqref{eq: SSA convex 12}~\cite{baysian_ssa}
\begin{equation}
\underset{\bb{X}}{\mathrm{minimize}} \frac{1}{2}\sum_{n=1}^N \left\|\bb{D}\bb{x}_n -\bb{s}_n\right\|_2^2 +\gamma_1\!\sum_{n=1}^N \|\bb{x}_n\|_1+\!\gamma_2\left\|\bb{X}\right\|_{2,1}, \label{eq: SSA convex 12 1}
\end{equation}
where $\gamma_1\geq0$ and $\gamma_2\geq0$ are the element-sparsity and row-sparsity regularization parameters, respectively.

In this paper, we extend the SSA problem to the convolutional sparse approximation (CSA) framework. Unlike its conventional counterpart, CSA allows local processing of large signals without first breaking them into vectorized overlapping blocks. Thus, it provides a global, single-valued, and shift-invariant model. Specifically, CSA uses a sum of convolutions instead of the matrix-vector product as in the standard sparse approximation model~\cite{CSC}. 

We first address the convolutional SSA (CSSA) problem with \textit{row-sparse} structure using the $\ell_{2,1}$-norm regularization (the convolutional extension of problem \eqref{eq: SSA convex 12}). Then, we discuss variations of the proposed method for solving problem \eqref{eq: SSA convex 12 1} and SSA with $\ell_{\infty,1}$-norm regularization in the CSA framework. We use the alternating direction method of multipliers (ADMM) as a base optimization approach for solving the corresponding problems. We investigate convolutional dictionary learning (CDL), and coupled feature learning in multimodal data based on CSSA. We evaluate the proposed CSSA and CDL algorithms by applying them to the multifocus image fusion and the near infrared (NIR) and visible light (VL) image fusion problems. Specifically, a novel NIR-VL image fusion method is proposed. MATLAB implementations of the proposed algorithms are available at~{\url{https://github.com/FarshadGVeshki/ConvSSA-IF}}.   

\section{Convolutional Simultaneous Sparse Approximation}
\label{sec: CSSA}
We aim to approximate the input signals $\bb{s}^{(n)} \in \mathbb{R}^{P},\; n=1,\dots,N,$ using the sparse feature maps with identical supports $\bb{X}^{(n)}\in \mathbb{R}^{P\times K},  \; n = 1,\dots,N$, and the dictionary $\bb{D}\in \mathbb{R}^{Q\times K}$. The columns of $\bb{X}^{(n)}$ and $\bb{D}$ are the convolutional SR elements and the convolutional filters, respectively. For simplicity, we consider the case where the input signals are one-dimensional arrays. The proposed method can be straightforwardly generalized to handling multi-dimensional arrays.  

\subsection{Problem Formulation}
The CSSA problem is formulated as follows
\begin{equation}
\begin{aligned}
\underset{ \{\bb{X}^{(n)}\}_{n=1}^N}{\mathrm{minimize}}  & \frac{1}{2}\sum_{n=1}^N \left\|\sum_{k=1}^K \bb{D}_k \ast \bb{X}_k^{(n)} \!-\!\bb{s}^{(n)}\right\|_2^2 \!+\!\lambda\! \sum_{n=1}^N\! \sum_{k=1}^K\!\left\|\bb{X}_k^{(n)}\right\|_{0}\\ 
\text{s.t.} \quad &\mathrm{Supp}\big(\bb{X}^{(m)}\big)=\mathrm{Supp}\big(\bb{X}^{(n)}\big),\quad m,n=1,\dots,N. \label{eq: SCSC}
\end{aligned}
\end{equation}
Using the $\ell_{2,1}$-norm\footnote{The mixed $\ell_{2,1,\dots,1}$-norm can be used for multi-dimensional input signal.}, a convex relaxation of \eqref{eq: SCSC} can be written as
\begin{equation}
\underset{ \{\bb{X}^{(n)}\}_{n=1}^N}{\mathrm{minimize}} \frac{1}{2}\sum_{n=1}^N \left\|\sum_{k=1}^K \bb{D}_k \ast \bb{X}_k^{(n)} -\bb{s}^{(n)}\right\|_2^2 \!+\!\lambda \sum_{k=1}^K\left\|\bb{\mathcal{X}}^{(k)}\right\|_{2,1} \label{eq: SCSC relaxed}
\end{equation}
where $\bb{\mathcal{X}}^{(k)}(p,\cdot) = [\bb{X}_k^{(1)}(p)\;\cdots\;\bb{X}_k^{(N)}(p)], \; p=1,\dots,P$.

\subsection{Optimization Procedure}
\label{sec: optimization}
The ADMM formulation of \eqref{eq: SCSC relaxed} can be written as
\begin{equation}
\begin{split}
\underset{ \{\bb{X}^{(n)}, \bb{Y}^{(n)}\}_{n=1}^N}{\mathrm{minimize}} &\frac{1}{2}\!\sum_{n=1}^N \!\left\|\sum_{k=1}^K\! \bb{D}_k \!\ast\! \bb{Y}_k^{(n)} \!-\!\bb{s}^{(n)}\right\|_2^2\! +\!\lambda\! \sum_{k=1}^K\!\left\|\bb{\mathcal{X}}^{(k)}\right\|_{2,1}\\
\text{s.t.} \quad &  \bb{X}^{(n)} 
= \bb{Y}^{(n)}, \; n=1,\cdots,N.\label{eq: SCSC ADMM}
\end{split}
\end{equation}
Then the ADMM iterations are given as 
\begin{align}
\begin{split}
&(\bb{Y}^{(n)})^{i+1} = \underset{ \bb{Y}^{(n)}}{\mathrm{argmin}} \frac{1}{2}\left\|\sum_{k=1}^K \bb{D}_k \ast \bb{Y}_k^{(n)} -\bb{s}^{(n)}\right\|_2^2 \\
& \hspace{1em}+ \frac{\rho}{2} \left\|\bb{Y}^{(n)} - (\bb{X}^{(n)})^i + (\bb{U}^{(n)})^i\right\|_{\rm F}^2, \; n = 1,\dots,N \label{eq: Y-update}
\end{split}\\
\begin{split}
&(\{\bb{X}^{(n)}\}_{n=1}^N)^{i+1} = \underset{ \{\bb{X}^{(n)}\}_{n=1}^N}{\mathrm{argmin}}\lambda \sum_{k=1}^K\left\|\bb{\mathcal{X}}^{(k)}\right\|_{2,1}\\
&\hspace{4em}+\frac{\rho}{2}\sum_{n=1}^N \left\|(\bb{Y}^{(n)})^{i+1} \!-\! \bb{X}^{(n)} \!+\! (\bb{U}^{(n)})^i\right\|_{\rm F}^2 \label{eq: X-update}
\end{split}\\
\begin{split}
&(\bb{U}^{(n)})^{i+1} \!=\! (\bb{Y}^{(n)})^{i+1} \!-\!(\bb{X}^{(n)})^{i+1} \!+\! (\bb{U}^{(n)})^i,n=1,\dots,N,\nonumber\\
\end{split}
\end{align}
where $\|\cdot\|_{\rm F}$ denotes the Frobenius norm of a matrix, $\{\bb{U}^{(n)}\}_{n=1}^N$ are the scaled Lagrangian multipliers, and $\rho>0$ is the ADMM penalty parameter. The $Y$-update step \eqref{eq: Y-update} entails $N$ convolutional regression subproblems which can be addressed using existing CSA methods (e.g.,~\cite{CSC}).

Since the $\ell_{2,1}$-norm is a separable sum of the $\ell_2$-norms of the rows, \eqref{eq: X-update} can be addressed in a row-wise manner using the proximal operator of the Euclidean norm. Using $\bb{W}^{(n)} = (\bb{Y}^{(n)})^{i+1} + (\bb{U}^{(n)})^i$, the solution to \eqref{eq: X-update} can be calculated as
\begin{multline}
\left([\bb{X}_k^{(1)}(p)\;\cdots\;\bb{X}_k^{(N)}(p)]\right)^{i+1} \\
=  \mathrm{prox}_{\frac{\lambda}{\rho}\|\cdot\|_2}\left([\bb{W}_k^{(1)}(p)\;\cdots\;\bb{W}_k^{(N)}(p)]\right),\\
k = 1,\dots,K, \; p = 1,\dots,P, \label{eq: prox2 step}
\end{multline}
with
\begin{equation}
\mathrm{prox}_{\tau\|\cdot\|_2}\big(\bb{a}\big) = \left(1-\frac{\tau}{\mathrm{max}(\|\bb{a}\|_2,\tau)} \right)\bb{a}. \label{eq: prox2}
\end{equation}
\subsection{Other Convex Formulations of CSSA}
Problem \eqref{eq: SCSC} can be alternatively relaxed using the $\ell_{\infty,1}$-norm. To address the resulting optimization problem, we only need to modify the $X$-update step of the ADMM algorithm explained in Subsection~\ref{sec: optimization}. Specifically, in \eqref{eq: prox2 step}, we need to replace $\mathrm{prox}_{\frac{\lambda}{\rho}\|\cdot\|_2}(\cdot)$ with the proximal operator of the $\ell_{\infty}$-norm $\mathrm{prox}_{\frac{\lambda}{\rho}\|\cdot\|_{\infty}}(\cdot)$, which is given as
\begin{equation}
\mathrm{prox}_{\tau\|\cdot\|_{\infty}}\big(\bb{a}\big) = \bb{a} - \tau {\Pi}_{(\|\cdot\|_1\leq1)}\left(\frac{\bb{a}}{\tau}\right), \label{eq: prox inf}
\end{equation}
where ${\Pi}_{(\|\cdot\|_1\leq1)}(\cdot)$ denotes the projection on the unit $\ell_1$-norm ball. Solving \eqref{eq: prox inf} requires iterative methods and it is more computationally expensive compared to computing \eqref{eq: prox2}.

The CSSA problem corresponding to \eqref{eq: SSA convex 12 1} can be written as
\begin{multline}
\underset{ \{\bb{X}^{(n)}\}_{n=1}^N}{\mathrm{minimize}} \; \frac{1}{2}\sum_{n=1}^N \left\|\sum_{k=1}^K \bb{D}_k \ast \bb{X}_k^{(n)} -\bb{s}^{(n)}\right\|_2^2 \\
+ \sum_{k=1}^K\left(\gamma_1\left\|\bb{\mathcal{X}}^{(k)}\right\|_{1,1}+\gamma_2\left\|\bb{\mathcal{X}}^{(k)}\right\|_{2,1}\right). \label{eq: SCSC 12 1}
\end{multline}
Problem \eqref{eq: SCSC 12 1} can be addressed using the method in Subsection~\ref{sec: optimization} after modifying the $X$-update step \eqref{eq: prox2 step} by replacing $\mathrm{prox}_{\frac{\lambda}{\rho}\|\cdot\|_2}(\cdot)$ with $\mathrm{prox}_{\frac{\gamma_1}{\rho}\|\cdot\|_1+\frac{\gamma_2}{\rho}\|\cdot\|_2}(\cdot)$, which can be calculated using
\begin{equation}
    \mathrm{prox}_{\tau\|\cdot\|_1+\kappa\|\cdot\|_2}(\bb{a}) =  \mathrm{prox}_{\kappa\|\cdot\|_2}\big(\mathcal{S}_{\tau}(\bb{a})\big),
\end{equation}
where the (elementwise) shrinkage operator $\mathcal{S}_{\tau}(a) = \mathrm{sign}(a) \mathrm{max}(0,|a|-\tau)$ is a proximal operator of the $\ell_1$-norm.

\section{Convolutional Dictionary Learning in Simultaneous Sparse Approximation Setup}
\label{sec: CDL}
Given $T$ sets of $N$ dependent input signals and their simultaneous SRs ($\{\bb{s}^{(t,n)}\}_{n=1}^N$ and $\{\bb{X}^{(t,n)}\}_{n=1}^N$, $t = 1,\dots,T$), the CDL problem can be formulated as follows
\begin{equation}
\begin{aligned}
\underset{\bb{D}}{\mathrm{minimize}} \; &\frac{1}{T}\sum_{t=1}^T \frac{1}{2}\sum_{n=1}^N  \Big\|\sum_{k=1}^K \bb{D}_k \ast \bb{X}_k^{(t,n)}-\bb{s}^{(t,n)}\Big\|_2^2 \\ 
\text{s.t.} \quad & \bb{D}_k, \in \bb{\mathcal{D}},\; k=1,\dots,K,\label{eq: SCDL}
\end{aligned}
\end{equation}
where $\bb{\mathcal{D}} = \left\{ \boldsymbol{d}\in \mathbb{R}^{Q} \ | \ \|\bb{d}\|_2 \leq1\right\}$. Problem \eqref{eq: SCDL} is a standard CDL problem and can be addressed using available batch~\cite{CSC} or online~\cite{online_CSC} CDL methods. Batch CDL requires all training data to be available at once, while online CDL is useful when the training samples are observed sequentially over time. Online CDL is also more computationally efficient when the total number of training samples (here $T\times N$) is larger than the number of filters in the dictionary (here $K$)~\cite{online_CSC}.

\subsection*{Convolutional Feature Learning in Multimodal Data}
If the input signals are multimodal and the order of modalities is fixed in all $T$ sets of training samples, we can extend the CDL problem \eqref{eq: SCDL} to learning multimodal convolutional dictionaries. This can be formulated as
\begin{equation}
\begin{aligned}
\underset{\{\bb{D}^{(n)}\}_{n=1}^N}{\mathrm{minimize}} \; &\frac{1}{T}\sum_{t=1}^T \frac{1}{2}\sum_{n=1}^N  \Big\|\sum_{k=1}^K \bb{D}^{(n)}_k \ast \bb{X}_k^{(t,n)}-\bb{s}^{(t,n)}\Big\|_2^2 \\ 
\text{s.t.} \quad & \bb{D}^{(n)}_k,  \in \bb{\mathcal{D}}, \; k=1,\dots,K,\; n=1,\dots,N,\label{eq: SCDL coupled}
\end{aligned}
\end{equation}
which can be addressed as $N$ separate CDL problems. Problem \eqref{eq: SCDL coupled} can be interpreted as learning correlated (coupled) features in multimodal data using the corresponding filters in the multimodal dictionaries $\{\bb{D}^{(n)}\}_{n=1}^N$. 

\section{NIR-VL Image Fusion based on CSSA}
The NIR images are characterized by high contrast resolutions, for example, in capturing vegetation scenes and imaging in low-visibility atmospheric conditions such as fog or haze~\cite{Tophat}. Based on these characteristics, the NIR images are used for enhancing outdoor VL images. In this section, we propose a NIR-VL image fusion method based on CSSA and CDL. The CSSA is performed using both $\ell_{1}$ and $\ell_{2,1}$ regularizations and also multimodal dictionaries. The steps of the proposed method for the fusion of a pair of NIR and VL images (denoted as $\bb{s}_{\rm n}$ and $\bb{s}_{\rm v}$, respectively) of the same sizes are explained as follows.

Since, the NIR images are presented in greyscale, they can be fused with the intensity components of the VL images which are usually available in the RGB (red-green-blue) format. Hence, in the first step, the VL image is converted to a color space (e.g., YCbCr), where the intensity (greyscale) component, denoted by $\bb{s}_{\rm v,g}$, is isolated from the color components of the image. Next, $\bb{s}_{\rm n}$ and $\bb{s}_{\rm v,g}$ are decomposed into their low-resolution components $\bb{s}_{\rm n}^{\rm b}$ and $\bb{s}_{\rm v,g}^{\rm b}$, and high-resolution components $\bb{s}_{\rm n}^{\rm h}$ and $\bb{s}_{\rm v,g}^{\rm h}$, for example, using lowpass filtering (more details are given in Subsection~\ref{sec: res NIR-VIS}). 

Using the proposed CSSA method and a pair of pre-learned multimodal NIR-VL dictionaries (denoted as $\bb{D}^{\rm n}$ and $\bb{D}^{\rm v}$), the convolutional SRs $\bb{X}^{\rm n}$ and $\bb{X}^{\rm v}$ are obtained for $\bb{s}_{\rm n}^{\rm h}$ and $\bb{s}_{\rm v,g}^{\rm h}$, respectively. The convolutional SRs are fused using the max-absolute-value fusion rule. This can be formulated as follows

\begin{equation}
\begin{split}
 \boldsymbol{F}_k^{\rm v}(i,j) \!=\!& \begin{cases}
    \boldsymbol{X}_k^{\rm v}(i,j), & \text{if $|\boldsymbol{X}_k^{\rm v}{(i,j)}|\geq|\boldsymbol{X}_k^{\rm n}(i,j)|$}\\
    0, & \text{otherwise}
  \end{cases}, \quad \\ 
 \boldsymbol{F}_k^{\rm n}(i,j) \!=\!& \begin{cases}
    \boldsymbol{X}_k^{\rm n}(i,j), & \text{if $|\boldsymbol{X}_k^{\rm n}(i,j)|>|\boldsymbol{X}_k^{\rm v}(i,j)|$}\\
    0, & \text{otherwise}
  \end{cases},
\end{split}
\end{equation}
where $\boldsymbol{F}_k^{\rm n}$ and $\boldsymbol{F}_k^{\rm v}$ are the fused convolutional SRs containing only the most significant representation coefficients at each entry. Moreover, the points $(i,j)$ represent the locations of all pixels in $\bb{s}_{\rm n}^{\rm h}$ and $\bb{s}_{\rm v,g}^{\rm h}$, $|\cdot|$ denotes the absolute value of a number, and $k = 1,\dots,K$ (number of filters in the dictionaries). The fused greyscale high-resolution component $\bb{s}_{\rm f,g}^{\rm h}$ is then reconstructed using 
\begin{equation*}
    \bb{s}_{\rm f,g}^{\rm h} = \sum_{k=1}^K \bb{F}_k^{\rm n} \ast \bb{D}_k^{\rm n} + \sum_{k=1}^K \bb{F}_k^{\rm v} \ast \bb{D}_k^{\rm v}.
\end{equation*}
The fused greyscale image $\bb{s}_{\rm f,g}$ is formed using $\bb{s}_{\rm f,g}^{\rm h}$  and the low-resolution component of the VL image
\begin{equation*}
\bb{s}_{\rm f,g} = \bb{s}_{\rm v,g}^{\rm b} + \bb{s}_{\rm f,g}^{\rm h}.
\end{equation*}
 Finally, the (YCbCr) image with $\bb{s}_{\rm f,g}$ as the intensity component and the color components of the VL image is converted back to the RGB format to form the fused color image $\bb{s}_{\rm f}$. 

\section{Experiments}
We first use the proposed CSSA methods with different sparsity structures for sparse approximation of a pair of NIR-VL images and compare the obtained SRs. Next, we use the proposed methods in multifocus and multimodal image fusion tasks and compare the results with existing image fusion methods. The convolutional dictionaries used in the experiments contain 32 filters of size $8\times 8$ and are learned using the online CDL method of \cite{online_CSC}. The training data consists of a NIR-VL image dataset and a multifocus image dataset, each containing 10 pairs of images. The NIR-VL and multifocus images are collected from the RGB-NIR Scene dataset~\cite{EPFL} and the Lytro dataset~\cite{Lytro}, respectively. The fusion results are evaluated both visually and based on objective evaluation metrics. Five metrics are used for objective evaluations: average entropy (EN), average peak signal-to-noise ratio (PSNR), the structural similarity index (SSIM)~\cite{SSIM}, spatial frequency (SF)~\cite{SF}, and edge intensity (EI)~\cite{EI}. 
\begin{figure}[h]
  \centering
  \subfloat[VL image]{\includegraphics[width=0.24\textwidth]{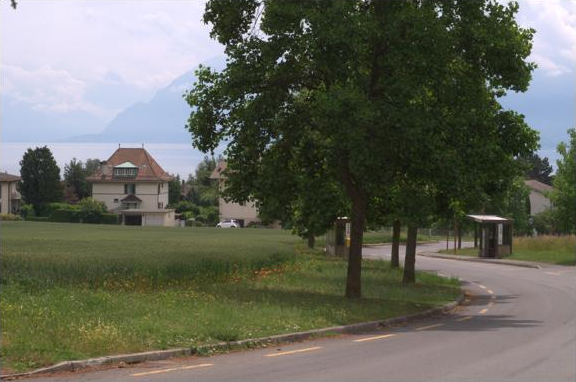}\label{fig:f11}} \
  \subfloat[NIR image]{\includegraphics[width=0.24\textwidth]{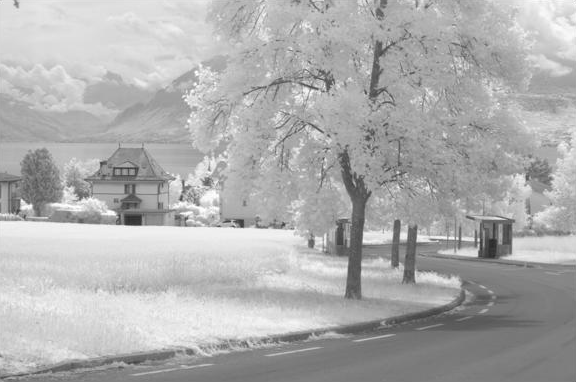}\label{fig:f12}}
  \caption{A pair of VL and NIR images.} \label{fig: VL-NIR}
\end{figure}

\begin{table*}[ht]\footnotesize{
\centering
\setlength{\tabcolsep}{1.7pt}
\renewcommand{\arraystretch}{0.6}
\begin{tabular}{lcccccccccccccccrr}
\toprule
& \multicolumn{3}{c}{\textbf{CSA}} & \; &\multicolumn{3}{c}{\textbf{CSSA-1}} & \quad\quad & \multicolumn{3}{c}{\textbf{CSSA-2a}} & \; & \multicolumn{3}{c}{\textbf{CSSA-2b}} & & \\
\cmidrule(lr){2-4} \cmidrule(lr){6-8} \cmidrule(lr){10-12} \cmidrule(lr){14-16}
$\lambda$ & Sparsity & Com. supp. & App. err. & & Sparsity & Com. supp. & App. err. & & Sparsity & Com. supp. & App. err. & & Sparsity & Com. supp. & App. err. & $\gamma_1$ & $\gamma_2$ \\
\cmidrule(){1-8} \cmidrule(){10-18}
$0.001$  & $0.0159$  & $2.92\%$ & $4.2326$   & & $0.0345$  & $100\%$ & $2.5245$   & & $0.0209$  & $33.52\%$ & $6.5948$   & & $0.0248$ & $32.91\%$ & $2.8497$   & $0.001$ & $0.001$\\
$0.01$   & $0.0094$  & $3.48\%$ & $40.2714$  & & $0.0183$  & $100\%$ & $36.7564$  & & $0.0165$  & $87.66\%$ & $40.9878$  & & $0.0201$ & $87.30\%$ & $24.9402$  & $0.001$ & $0.01$ \\
$0.05$   & $0.0038$  & $4.22\%$ & $148.9280$ & & $0.0073$  & $100\%$ & $137.0919$ & & $0.0091$  & $37.61\%$ & $72.8397$  & & $0.0110$ & $35.31\%$ & $51.6590$  & $0.01$  & $0.01$\\
$0.1$    & $0.0020$  & $4.34\%$ & $241.0831$ & & $0.0040$  & $100\%$ & $221.8418$ & & $0.0040$  & $98.57\%$ & $223.7260$ & & $0.0045$ & $98.74\%$ & $214.8775$ & $0.001$ & $0.1$\\
$0.5$    & $0.0001$  & $1.14\%$ & $657.2492$ & & $0.0004$  & $100\%$ & $625.3224$ & & $0.0034$  & $87.96\%$ & $240.1183$ & & $0.0038$ & $87.77\%$ & $233.0870$ & $0.01$  & $0.1$\\
\bottomrule 
\end{tabular}
\caption{Comparison of the convolutional SRs of the multimodal images in Fig.~\ref{fig: VL-NIR} obtained using the (unstructured) CSA method of~\cite{CSC}, the proposed CSSA method with $\ell_{2,1}$ regularization (CSSA-1), and the proposed CSSA method with $\ell_{2,1}$ and $\ell_{1}$ regularizations using a single dictionary (CSSA-2a) and two (multimodal) dictionaries (CSSA-2b) in terms of ratio of the nonzero entries (sparsity), the percentage of overlapping nonzero entries (Com. supp.), and the residuals power (App. err.). The convolutional dictionaries used consist of 32 filters of size $8\times 8$ and are learned using a set of 10 pairs of NIR-VL images.} \label{table: 1}}
\end{table*}

\subsection{Performance Comparison}
We investigate the performances of the proposed CSSA methods in capturing the underlying structures of the NIR-VL images in Fig.~\ref{fig: VL-NIR} in terms of sparsity, the overlap between supports of the SRs, and the residual power. We compare the results also with those obtained using the unstructured CSA method of~\cite{CSC}. The results obtained using different values of the sparsity regularization parameters are summarized in Table~\ref{table: 1}. As can be seen, the unstructured CSA leads to inconsiderable overlaps between the supports of the convolutional SRs, indicating the fact that CSA with no structure cannot effectively capture the existing correlations between the input images. The CSSA method with $\ell_{2,1}$ regularization (CSSA-1) results in convolutional SRs with identical supports ($100\%$ overlap). However, the imposed structure leads to lower sparsity in the SRs and higher approximation errors.

CSSA using $\ell_{1}$ and $\ell_{2,1}$ regularizations (CSSA-2a) allows to relax the identical supports constraint. Specifically, the use of a larger element-sparsity parameter $\gamma_1$ allows for a smaller overlap between the supports of the SRs. This approximation model is more appropriate when the correlated input signals can contain (or lack) specific features. For example, in NIR-VL images, some details are visible only in one of the input images. This model can be also extended to learn the nonlinear local relationships in the multimodal data in terms of a set of multimodal dictionaries (CSSA-2b). The results in Table~\ref{table: 1} show that the use of multimodal dictionaries leads to considerably more accurate approximations while achieving SRs with the same level of sparsity compared to the case where a single dictionary is used for both modalities. 
\begin{figure}[ht]
  \centering
  \subfloat[The method of \cite{Tophat}]{\includegraphics[width=0.48\textwidth]{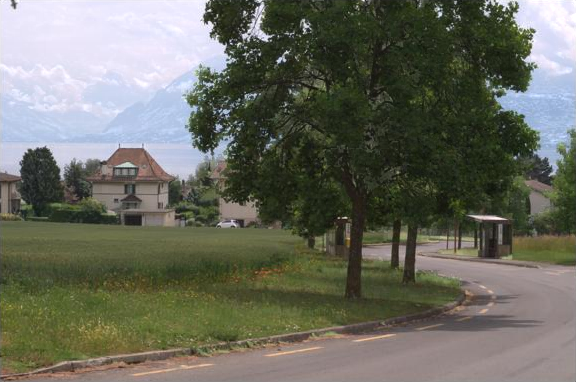}\label{fig:f13}}\\ 
  \subfloat[CSSA]{\includegraphics[width=0.48\textwidth]{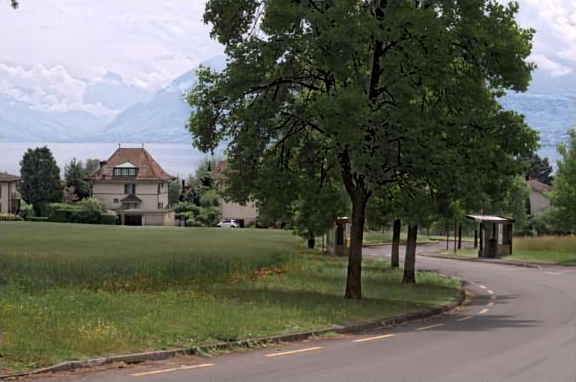}\label{fig:f14}}
  \caption{Visible light and near infrared image fusion results.} \label{fig: MM}
\end{figure}

\subsection{NIR-VL Image Fusion Results}
\label{sec: res NIR-VIS}
We benchmark the performance of the proposed NIR-VL image fusion method by comparing our results with those obtained using the fusion method of~\cite{Tophat}. There are $51$ pairs of outdoor NIR-VL images labeled as ``\textit{country}" in the RGB-NIR Scene dataset. We use 10 pairs of these images for CDL, and the remainder 41 images are used as the test dataset. The CSSA is performed using parameters $\rho=10$, $\gamma_1 = 0.001$ and $\gamma_2 = 0.01$. The lowpass filtering is performed using the \textit{lowpass} function from the SPORCO library~\cite{wohlberg-2016-sporco} with the regularization parameter of 5.

Fig.~\ref{fig: MM} shows the fusion results for the NIR-VL images in Fig.~\ref{fig: VL-NIR}. The average objective evaluation results obtained for the entire test dataset are reported in Table~\ref{table: objective evaluation ressults}. As it can be seen in Fig.~\ref{fig: MM}, the proposed fusion method achieves higher contrast resolutions, which is also reflected in larger entropy, spatial frequency, and edge intensity values in Table~\ref{table: objective evaluation ressults}. However, method of \cite{Tophat} results in better SSIM and PSNR. This can be explained by the fact that in the proposed method, the fused images are reconstructed from sparse approximations, while the original pixel values are used in \cite{Tophat}. 

\begin{table}[htb]\footnotesize{
\centering
\setlength{\tabcolsep}{1.7pt}
\renewcommand{\arraystretch}{0.6}
\begin{tabular}{lcccccc}
\toprule
 & & \multicolumn{2}{c}{\textbf{Multifocus}} & & \multicolumn{2}{c}{\textbf{NIR-VL}} \\
\cmidrule(lr){3-4} \cmidrule(lr){6-7}
\textbf{Metrics}& & The method of~\cite{CSCfusion}   & CSSA  &  & The method of~\cite{Tophat}      & CSSA\\
\midrule
$\mathrm{EN}$     &  & $7.4273$        & $\bb{7.4371}$  &  & $7.0630$        & $\bb{7.2649}$  \\
$\mathrm{SF}$      &  & $16.6709$       & $\bb{16.8536}$ &  & $18.1657$       & $\bb{20.0136}$ \\
$\mathrm{EI}$      &  & $60.6051$       & $\bb{61.9919}$ &  & $59.7521$       & $\bb{73.0752}$ \\
$\mathrm{SSIM}$    &  & $0.8491$        & $\bb{0.8498}$  &  & $\bb{0.7629}$   & $0.7574$  \\
$\mathrm{PSNR}$    &  & $\bb{27.8952}$  & $27.5893$      &  & $\bb{20.3470}$ & $19.3590$ \\ \bottomrule
\end{tabular}
\caption{Average objective evaluation results using different methods. The best results are shown in bold.} \label{table: objective evaluation ressults}}
\end{table}

\subsection{Multifocus Image Fusion}
\begin{figure}[ht]
  \centering
  \subfloat[Input 1]{\includegraphics[width=0.14\textwidth]{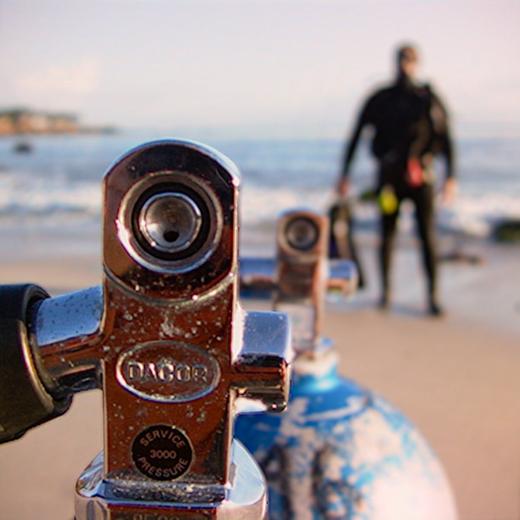}\label{fig:f31}} \
  \subfloat[Input 2]{\includegraphics[width=0.14\textwidth]{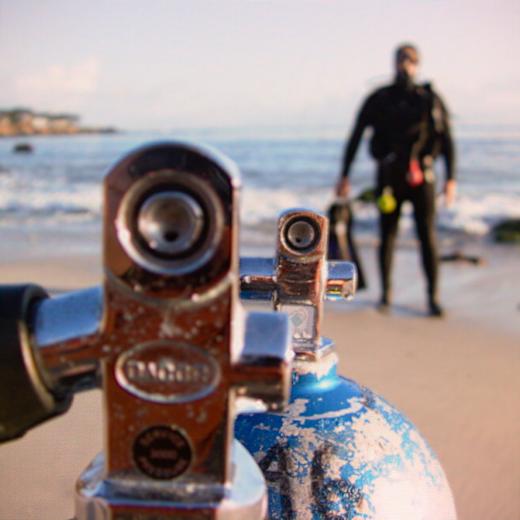}\label{fig:f32}} \
  \subfloat[Input 3]{\includegraphics[width=0.14\textwidth]{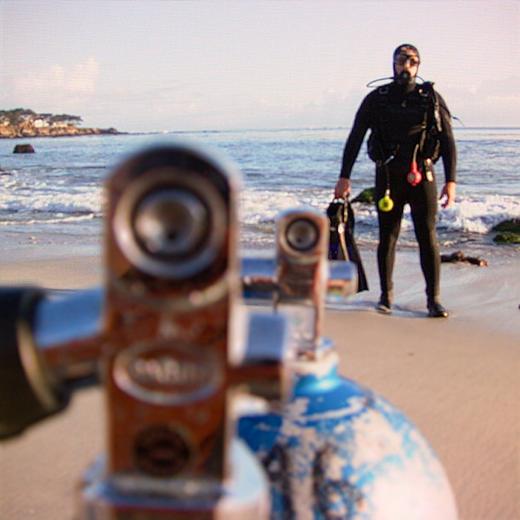}\label{fig:f33}}\\
   \subfloat[The method of \cite{CSCfusion}]{\includegraphics[width=0.22\textwidth]{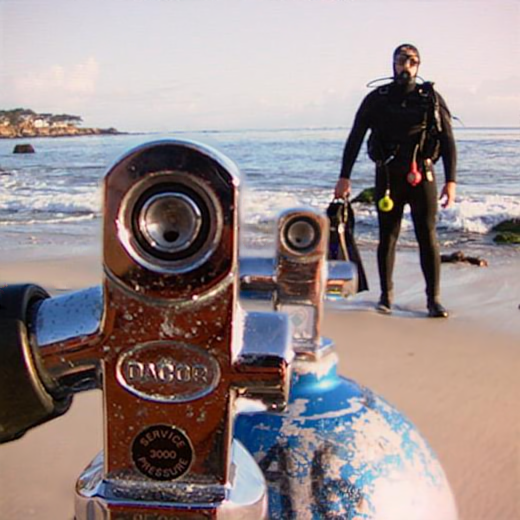}\label{fig:f34}}\ 
    \subfloat[CSSA]{\includegraphics[width=0.22\textwidth]{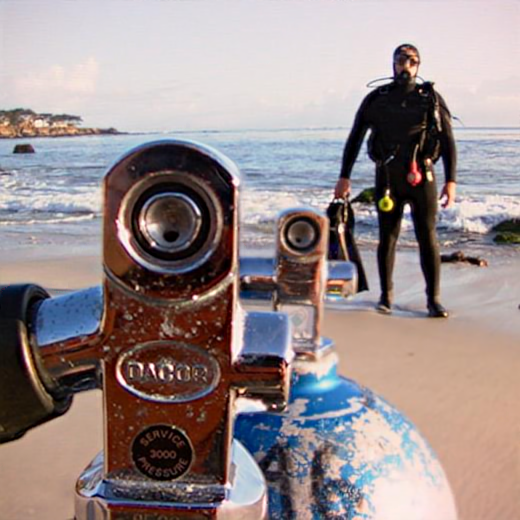}\label{fig:f35}}
  \caption{Multifocus image fusion results.} \label{fig: MF 3inputs}
\end{figure}

In this section, we modify the multifocus image fusion method of \cite{CSCfusion} to incorporate CSSA instead of using unconstrained CSA and compare the resulting performances. The test dataset contains 10 pairs of multifocus images (different from the training dataset) and 4 sets of triple multifocus images. The CSSA is performed using only the $\ell_1$-norm regularization with parameters $\rho = 10$ and $\lambda = 0.01$. The method of \cite{CSCfusion} uses the max-$\ell_1$-norm rule for fusing the convolutional SRs. In the modified fusion method, we fuse the convolutional SRs (with identical supports) using the elementwise maximum absolute value rule to generate the fused convolutional SRs. All other steps of the two algorithms are identical. The obtained fusion results show that the use of CSSA leads to considerable improvements in terms of higher contrast resolutions and better fusion of multifocus edges (boundaries where one side is in-focus and the other side is out of focus). Fig.~\ref{fig: MF 3inputs} shows an example of fusion results obtained using the two methods. The objective evaluation results in Table~\ref{table: objective evaluation ressults} also indicate that CSSA improves on the overall performance of the CSA-based multifocus image fusion method of \cite{CSCfusion} . 


\section{Conclusion}
Algorithms for convolutional simultaneous sparse approximation with different sparsity structures based on the alternating direction method of multipliers have been proposed. We have evaluated the effectiveness of the proposed methods by using them in two different categories of image fusion problems and compared the obtained results with those of existing image fusion methods. In particular, a novel near infrared and visible light image fusion method based on convolutional simultaneous sparse approximation has been proposed.

\end{document}